\documentclass[11pt,a4paper]{article}
\usepackage[nohyperref]{emnlp18/emnlp2018}
\usepackage{times}
\usepackage{latexsym}
\usepackage{url} 
\usepackage{graphicx}  
\usepackage{microtype}
\usepackage{bbm}

\usepackage{subcaption}
\usepackage{comment}

\usepackage{amsmath}
\usepackage{amssymb}
\aclfinalcopy
\usepackage{capt-of}

\title{Understanding Deep Learning Performance through an Examination of Test Set Difficulty: A Psychometric Case Study}

\author{John P. Lalor$^{1*}$, Hao Wu$^2$, Tsendsuren Munkhdalai$^3$, Hong Yu$^{1,4}$ \\
  $^1$College of Information and Computer Sciences, University of Massachusetts, Amherst\\
  $^2$ Department of Psychology and Human Development, Vanderbilt University\\
  $^3$Microsoft Research, Montr\'{e}al, Qu\'{e}bec \\
  $^4$Department of Computer Science, University of Massachusetts, Lowell\\
    {*\tt lalor@cs.umass.edu} 
}

\date{}
\begin{document}
\maketitle

\begin{abstract}
  Interpreting the performance of deep learning models beyond test set accuracy is challenging.
  Characteristics of individual data points are often not considered during evaluation, and each data point is treated equally.
  We examine the impact of a test set question's difficulty to determine if there is a relationship between difficulty and performance.
  We model difficulty using well-studied psychometric methods on human response patterns.
  Experiments on Natural Language Inference (NLI) and Sentiment Analysis (SA) show that the likelihood of answering a question correctly is impacted by the question's difficulty. 
  As DNNs are trained with more data, easy examples are learned more quickly than hard examples.
\end{abstract}

\section{Introduction}

One method for interpreting deep neural networks (DNNs) is to examine model predictions for specific input examples, e.g. testing for shape bias as in~\citet{ritter2017cognitive}.
In the traditional classification task, the difficulty of the test set examples is not taken into account.
The number of correctly-labeled examples is tallied up and reported.
However, we hypothesize that it may be worthwhile to use difficulty when evaluating DNNs.
For example, what does it mean if a trained model answers the more difficult examples correctly, but cannot correctly classify what are seemingly simple cases?
Recent work has shown that for NLP tasks such as Natural Language Inference (NLI), models can achieve strong results by simply using the hypothesis of a premise-hypothesis pair and ignoring the premise entirely \cite{gururangan2018annotation,TSUCHIYA18.786,poliak2018hypothesis}.

In this work we consider understanding DNNs by looking at the difficulty of specific test set examples and comparing DNN performance under different training scenarios.
Do DNN models learn examples of varying difficulty at different rates?
If a model does well on hard examples and poor on easy examples, then can we say that it has really learned anything?
In contrast, if a model does well on easy items, because a dataset is all easy, have we really ``solved'' anything?

To model difficulty we use Item Response Theory (IRT) from psychometrics~\citep{baker_item_2004}.
IRT models characteristics such as difficulty and discrimination ability of specific examples (called ``items''\footnote{For the remainder of the paper we will refer to a single test set example as an ``item'' for consistency.}) in order to estimate a latent ability trait of test-takers.
Here we use IRT to model the difficulty of test items to determine how DNNs learn items of varying difficulty. 
IRT provides a well-studied methodology for modeling item difficulty as opposed to more heuristic-based difficulty estimates such as sentence length.
IRT was previously used to build a new test set for the NLI task~\cite{lalor2016beyond} and show that model performance is dependent on test set difficulty.
In this work we use IRT to probe specific items to try to analyze model performance at a more fine-grained level, and expand the analysis to include the task of SA.

We train three DNNs models with varying training set sizes to compare performance on two NLP tasks: NLI and Sentiment Analysis (SA).
Our experiments show that a DNN model's likelihood of classifying an item correctly is dependent on the item's difficulty.
In addition, as the models are trained with more data, the odds of answering easy examples correctly increases at a faster rate than the odds of answering a difficult example correctly.
That is, performance starts to look more human, in the sense that humans learn easy items faster than they learn hard items.

That the DNNs are better at easy items than hard items seems intuitive but is a surprising and interesting result since the item difficulties are modeled \textit{from human data}.
There is no underlying reason that the DNNs would find items that are easy for humans inherently easy.
To our knowledge this is the first work to use a grounded measure of difficulty learned from human responses to understand DNN performance.
Our contributions are as follows: (i) we use a well-studied methodology, IRT, to estimate item difficulty in two NLP tasks and show that this human-estimated difficulty is a useful predictor of DNN model performance, (ii) we show that as training size increases DNN performance trends towards expected human performance.\footnote{Code and data available at \tt{http://jplalor.github.io}}

\section{Methods}

\begin{table*}[t]
	\small
	\begin{tabular}{p{6.5cm}p{5.5cm}ll}
		\hline \bf Premise & \bf Hypothesis & \bf Label  & \bf Difficulty\\ \hline
		A little girl eating a sucker & A child eating candy & Entailment & -2.74 \\
		People were watching the tournament in the stadium & The people are sitting outside on the grass & Contradiction & 0.51 \\
		Two girls on a bridge dancing with the city skyline in the background & The girls are sisters. & Neutral & -1.92\\ 
		Nine men wearing tuxedos sing & Nine women wearing dresses sing & Contradiction  & 0.08 \\ \hline
	\end{tabular}
	\caption{Examples of sentence pairs from the SNLI data sets, their corresponding gold-standard label, and difficulty parameter ($b_i$) as measured by IRT (\S \protect{\ref{ssec:difficulty}}).}
	\label{tab:examples_rte}
\end{table*}

\begin{table*}[t]
	\small
	\begin{tabular}{p{13cm}ll}
		\hline \bf Phrase & \bf Label  & \bf Difficulty\\ \hline
		The stupidest, most insulting movie of 2002's first quarter. & Negative & -2.46 \\
		Still, it gets the job done - a sleepy afternoon rental. & Negative & 1.78 \\
		An endlessly fascinating, landmark movie that is as bold as anything the cinema has seen in years. & Positive & -2.27\\
		Perhaps no picture ever made has more literally showed that the road to hell is paved with good intentions. & Positive  & 2.05 \\ \hline
	\end{tabular}
	\caption{Examples of phrases from the SSTB data set, their corresponding gold-standard label, and difficulty parameter ($b_i$) as measured by IRT (\S \protect{\ref{ssec:difficulty}}).}
	\label{tab:examples_sent}
\end{table*}

\subsection{Estimating Item Difficulty}
\label{ssec:difficulty}

To model item difficulty we use the Three Parameter Logistic (3PL) model from IRT~\citep{baker2001basics,baker_item_2004,lalor2016beyond}.
The 3PL model in IRT models an individual's latent ability ($\theta$) on a task as a function of three item characteristics: discrimination ability ($a$), difficulty ($b$), and guessing ($c$).
For a particular item $i$, the probability that an individual $j$ will answer item $i$ correctly is a function of the individual's ability and the three item characteristics:

\begin{equation} 
p_{ij}(\theta_j) = c_i + \frac{1 - c_i}{1 + e^{-a_i(\theta_j - b_i)}}
\end{equation}

where $a_i$ is the discrimination parameter (the value of the function slope at it's steepest point), $b_i$ is the difficulty parameter (the value where $p_{ij}(\theta_j) = 0.5)$, and $c_i$ is the guessing parameter (the lower asymptote of the function).
%
For a set of items $I$ and a set of individuals $J$, the likelihood of each individual in $J$'s responses to the items in $I$ is:

\begin{equation}
L = \prod_{j=1}^J \prod_{i=1}^I p_{ij}(\theta_j)^{y_{ij}} q_{ij}(\theta_j)^{(1-y_{ij})}
\end{equation}

where $q_{ij}(\theta_j) = 1 - p_{ij}(\theta_j)$ and $y_{ij} = 1$ if individual $j$ answered item $i$ correctly and $y_{ij}=0$ otherwise.
Item parameters and individual ability are jointly estimated from a set of individuals' response patterns using an Expectation-Maximization algorithm~\citep{bock1981marginal}.

In this work we focus on the difficulty parameter $b_i$, which represents the latent ability level at which an individual has a 50\% chance of answering item $i$ correctly.
Low values of $b_i$ are associated with easier items (since an individual with low ability has a 50\% chance of answering correctly), and higher values of $b_i$ represent more difficult items.

\begin{figure*}[th]
	\centering
	\includegraphics[width=\linewidth]{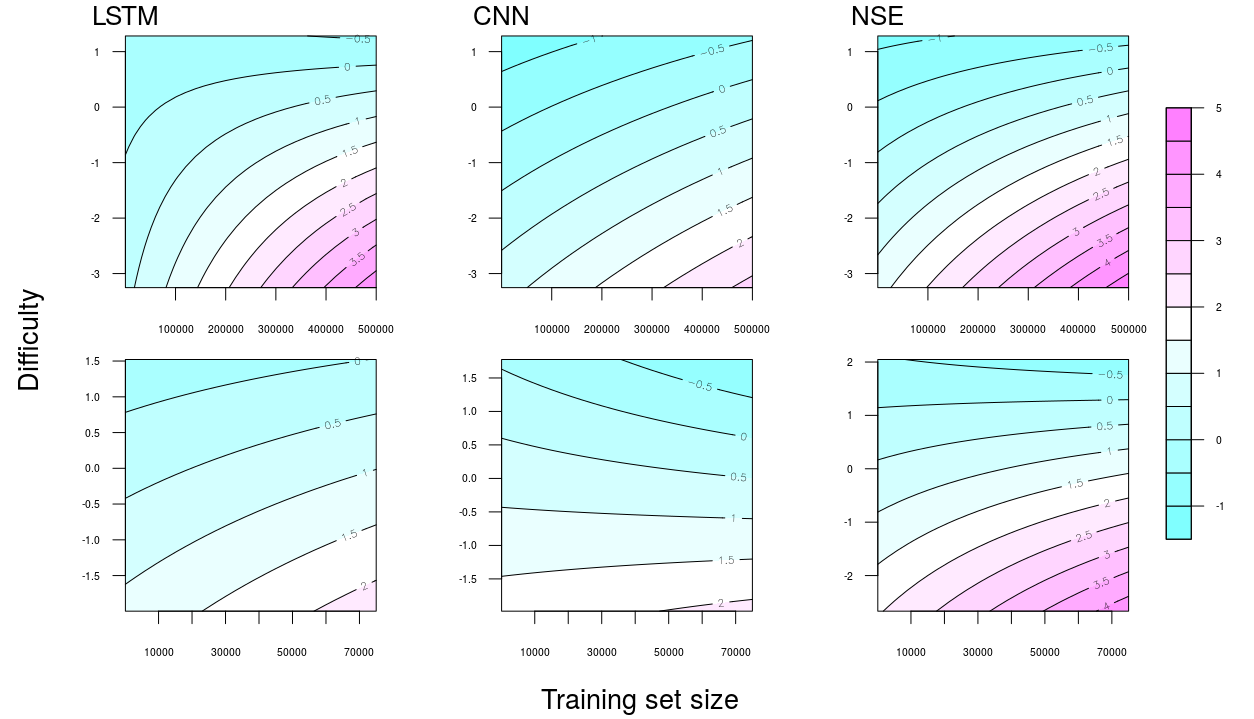}
	\caption{Contour plots showing log-odds of labeling an item correctly for NLI (top row) and SA (bottom row) as a function of training set size (x-axis) and item difficulty (y-axis). Each line in the plots represents a single log-odds value for labeling an item correctly. Blue indicates low log-odds of labeling an item correctly, and pink indicates high log-odds of labeling an item correctly. The contour colors are consistent across plots and log-odds values are shown in the legend on the right.}
	\label{fig:contours}
\end{figure*}

\subsection{Data}

\begin{table}[t]
\small
\centering
\begin{tabular}{|lc|}
  \hline \bf Dataset &  \bf Fleiss' $\kappa$  \\ \hline
  SNLI 4GS Contradiction & 0.37\\
  SNLI 4GS Entailment & 0.48\\
  SNLI 4GS Neutral & 0.41\\
  SNLI 5GS Contradiction & 0.59\\
  SNLI 5GS Entailment & 0.63\\
  SNLI 5GS Neutral & 0.54\\
  Sentiment Analysis & 0.52\\
\hline
\end{tabular}
\caption{Fleiss' $\kappa$ scores for the NLI and SA annotations collected from AMT. }\label{tab:irr}
\end{table}

To estimate item difficulties for NLI, we used the pre-trained IRT models of~\citet{lalor2016beyond} and extracted the difficulty item parameters.
The data consists of approximately 1000 human annotator responses from Amazon Mechanical Turk (AMT) for a selection of 180 premise-hypothesis pairs from the SNLI data set~\citep{bowman_large_2015}.
Each AMT worker (Turker) was shown the premise-hypothesis pairs and was asked to indicate whether, if the premise was taken to be true, the hypothesis was (a) definitely true (\textit{entailment}), (b) maybe true (\textit{neutral}), or (c) definitely not true (\textit{contradiction}). 


For SA, we collected a new data set of labels for 134 examples randomly selected from the Stanford Sentiment Treebank (SSTB) \cite{socher2013recursive}, using a similar AMT setup as~\citet{lalor2016beyond}.
For each randomly selected example, we had 1000 Turkers label the sentence as very negative, negative, neutral, positive, or very positive. 
We converted these responses to binary positive/negative labels and fit a new IRT 3PL model (\S \ref{ssec:difficulty}) using the \textit{mirt} R package~\citep{chalmers_mirt:_2015}.
Very negative and negative labels were binned together, and neutral, positive, and very positive were binned together.

Tables \ref{tab:examples_rte} and \ref{tab:examples_sent} show examples of the items in our data sets, and the difficulty values estimated from the IRT models.
The first example in Table \ref{tab:examples_rte} is a clear case of \textit{entailment}, where if we assume that the premise is true, we can infer that the hypothesis is also true.
The label of the second example in SNLI is \textit{contradiction}, but in this case the result is not as clear.
There are sports stadiums that offer lawn seating, and therefore this could potentially be a case of entailment (or neutral).
Either way, one could argue that the second example here is more difficult than the first.
Similarly, the first two examples of Table \ref{tab:examples_sent} are interesting.
Both of these items are labeled as \textit{negative} examples in the data set.
The first example is clear, but the second one is more ambiguous.
It could be considered a mild complement, since the author still endorses renting the movie.
Therefore you could argue again that the second example is more difficult than the first.
The learned difficulty parameters reflect this difference in difficulty in both cases.

Inter-rater reliability scores for the collected annotations are showin in Table \ref{tab:irr}.
Scores for the NLI annotations were calculated when the original dataset was collected and are reproduced here \citep{lalor2016beyond}.
Human annotations for the SA annotations were converted to binary before calculating the agreement. 
We see that the agreement scores are in the range of 0.4 to 0.6 which is considered moderate agreement \citep{landis1977measurement}.
With the large number of annotators it is to be expected that there is some disagreement in the labels.
However this disagreement can be interpreted as varying difficulty of the items, which is what we expect when we fit the IRT models.

\subsection{Experiments}
Our goal in this work is to understand how DNN performance on items of varying difficulty changes under different training scenarios.
To test this, we trained three DNN models using subsets of the original SNLI and SSTB training data sets: (i) Long Short Term Memory Network (LSTM)~\citep{bowman_large_2015}, (ii) Convolutional Neural Network (CNN)~\citep{D14-1181}, and (iii) Neural Semantic Encoder (NSE), a type of memory-augmented RNN~\citep{munkhdalai2016neural}.\footnote{Please refer to the appendix for model details.}
For each task (NLI and SA), we randomly sampled subsets of training data, from 100 examples up to and including the full training data sets.\footnote{We sampled 100, 1000, 2000, 5000, 10000, 50000, 100000, 200000, and 500000 examples for NLI, and sampled 100, 1000, 5000, 10000, 50000, and 75000 examples for SA.}
We trained each model on the training data subsets, using the original development sets for early stopping to prevent overfitting.
The IRT data with difficulty estimates were used as test sets for the trained models.

Once the models were trained and had classified the IRT data sets, we fit logistic regression models to predict whether a DNN model would label an item correctly, using the training set size and item difficulty as the dependent parameters.

\section{Results}

Figure \ref{fig:contours} plots the contour plots of our learned regression models.
The top row plots results for the NLI task, and the bottom row plots results for the SA task.
From left to right in both rows, the plots show results for the LSTM, CNN, and NSE models.
In each plot, the x-axis is the training set size, the y-axis is the item difficulty, and the contour lines represent the log-odds that the DNN model would classify an item correctly.
As the plots show, item difficulty has a clear effect on classification.
Easier items have higher odds of being classified correctly across all of the training set sizes.
In addition, the slopes of the contour lines are steeper at lower levels of difficulty.
This indicates that, moving left to right along the x-axis, a model's odds of answering an easy item correctly increase more quickly than the odds of answering a harder item correctly.

The contour plots for the CNN and NSE models on the SA task (Figure \ref{fig:contours}, second row middle and right plots) show that the easier items have higher likelihood of being classified correctly, but the odds for the most difficult items decrease as training size increases.
This suggests that these models are learning in such a way that improves performance on easy items but has a negative effect on hard items.
This result is important for interpretability, as it could inform stakeholder decisions if they need to have difficult examples classified.

The idea that easy items should be easier than hard items is consistent with learning strategies in humans.
For example, when teaching new concepts to students, easier concepts are presented first so that the students can learn patterns and core information before moving to more difficult concepts~\cite{collins1988cognitive,arroyo2010effort}.
As students do more examples, all questions get easier, but easy questions get easier at a faster rate.
Our result is also consistent with the key assumptions of curriculum learning~\cite{bengio2009curriculum}.

\section{Related Work}

\citet{lalor2016beyond} introduced the idea of applying IRT evaluation to NLP tasks.
They built a set of scales using IRT for NLI and evaluated a single LSTM neural network to demonstrate the effectiveness of the evaluation, but did not evaluate other NLP models or tasks. 
\citet{conf/ecai/Martinez-Plumed16} consider IRT in the context of evaluating ML models, but they do not use a human population to calibrate the models, and obtain results that are difficult to interpret under IRT assumptions.

There has been work in the NLP community around modeling latent characteristics of data~\cite{bruce1999recognizing} and annotators~\citep{hovy2013learning}, but none that apply the resulting metrics to interpret DNN models.
\citet{passonneau2014benefits} model the probability a label is correct with the probability of an annotator to label an item correctly according to the~\citet{dawid1979maximum} model, but do not consider difficulty or discriminatory ability of the data points.

One-shot learning is an attempt to build ML models that can generalize after being trained on one or a few examples of a class as opposed to a large training set~\citep{lake2013one}.
One-shot learning attempts to mimic human \textit{learning behaviors} (i.e., generalization after being exposed to a small number of training examples)~\citep{lake2016}.
Our work instead looks at comparisons to human \textit{performance}, where any learning (on the part of models) has been completed beforehand.
Our goal is to analyze DNN models and training set variations as they affect ability in the context of IRT.

\section{Discussion}

In this work we have shown that DNN model performance is affected by item difficulty as well as training set size.
This is the first work that has used a well-established method for estimating difficulty to analyze DNN model performance as opposed to heuristics.
DNN models perform better on easy items, and as more data is introduced in training, easy items are learned more quickly than hard items.
Learning easy examples faster than harder examples is what would be expected when examining human response patterns as they learn more about a subject.
However this has not previously been shown to be true in DNN models.

That the results are consistent across NLI and SA shows that the methods can be applied to a number of NLP tasks.
The SA results do show that the odds of labeling a difficult item correctly decrease with more training data \ref{fig:contours}.
It could be the case that these difficult items in the SA task are more subjective than the easier items, for example a review that is fairly neutral and is split between positive and negative annotations.
These cases would be more difficult for a model to label, and are worth examining in more detail.
By identifying items such as these as difficult makes it easier to see where the model is going wrong and allows for research on better way to represent these cases.

This result has implications for how machine learning models are evaluated across tasks.
The traditional assumption that the test data is drawn from the same distribution as the training data, makes it difficult to understand how a model will perform in settings where that assumption does not hold.
However, if the difficulty of test set data is known, we can better understand what kind of examples a given model performs well on, and specific instances where a model underperforms (e.g. the most difficult examples).
In addition, researhers can build test sets that consist of a specific type of data (very easy, very hard, or a mix) to evaluate a trained model under specific assumptions to test generalization ability in a controlled way.
This could allow for more confidence in model performance in more varied deployment settings, since there would be a set of tests a model would have to pass before being deployed.

It is important to note that the difficulty parameters were estimated \textit{from a human population}, meaning that those items that are difficult for humans are in fact more difficult for the DNN models as well.
This does not need to be the case given that DNNs learn very different patterns, etc. than humans.
In fact there were exceptions in our results which shows that these models should be carefully examined using techniques like those described here.
Future work can investigate why this is the case and how we can leverage this information to improve model performance and interpretability.
\section*{Acknowledgments}
We thank the AMT Turkers who completed our annotation task.
This work was supported in part by the HSR\&D award IIR 1I01HX001457 from the United States Department of Veterans Affairs (VA).
We also acknowledge the support of LM012817 from the National Institutes of Health.
This work was also supported in part by the Center for Intelligent Information Retrieval.
The contents of this paper do not represent the views of CIIR, NIH, VA, or the United States Government.


\bibliography{jlalor}
\bibliographystyle{emnlp18/acl_natbib_nourl}

\appendix

\section{DNN Model Architecture}

Here we provide a brief overview of the model architectures for the deep neural network (DNN) models used in our experiments. For additional details we refer the reader to the original papers.

\subsection{Long Short Term Memory}
The Long Short Term Memory (LSTM) model used here was provided by \cite{bowman_large_2015} with the release of the SNLI corpus.
The model consists of two LSTM sequence-embedding models \cite{hochreiter_long_1997}, one to encode the premise and another to encode the hypothesis.
The two sentence encodings are then concatenated and passed through three tanh layers.
Finally, the output is passed to a softmax classifier layer to output probabilities over the task classes.
For SA, we kept the same architecture but used a single LSTM layer to encode the input text.

\subsection{Convolutional Neural Network}
We used the convolutional neural network (CNN) model of \cite{D14-1181} in our experiments.
For each input, the word vector representation of the input tokens were concatenated together to form a matrix.
A series of convolutional operations were applied, followed by a max-pooling operation and a fully connected softmax classifier layer.
More concretely, for an input sentence $\mathbf{x}$, let $\mathbf{x}_i$ be the word vector representation of the $i$-th word in $\mathbf{x}$.
The convolution operation of filter $\mathbf{w}$ over a window of length $h$ starting with word $\mathbf{x}_i$ results in a context vector $c_i$:

\begin{equation}
c_i = f(\mathbf{w} \cdot \mathbf{x}_{i:i+h-1} + b)
\end{equation}

where $b$ is a bias term \cite{D14-1181}.
The filter is applied over all windows in the sentence to generate a feature-map, and max-pooling is used to identify the feature for this particular filter.
The process is repeated with multiple filters, and the output features are then passed to a softmax classification layer to output probabilities over the class labels \cite{D14-1181}.
For NLI, the premise and hypothesis sentences were concatenated before encoding.

\subsection{Neural Semantic Encoder}
Neural Semantic Encoder (NSE) is a memory-augmented neural network that uses \textit{read}, \textit{compose}, and \textit{write} operations to evolve and maintain an external memory \cite{munkhdalai2016neural}:

\begin{align}
o_t &= f_r^{LSTM}(x_t) \\
z_t &= softmax(o_t^\top M_{t-1}) \\
m_{r,t} &= z_t^\top M_{t-1} \\
c_t &= f_c^{MLP} (o_t, m_{r,t}) \\
h_t &= f_w^{LSTM}(c_t) \\
M_t = M_{t-1} (\mathbf{1} &- (z_t \otimes e_k)^\top ) + (h_t \otimes e_l)(z_t \otimes e_k)^\top
\end{align}

where $f_r^{LSTM}$ is the read function, $f_c^{MLP}$ is the composition function, $f_w^{LSTM}$ is the write function, $M_t$ is the external memory at time $t$, and $e_l \in R^l$ and $e_k \in R^k$ are vectors of ones \cite{munkhdalai2016neural}.

For NLI, the premise and hypothesis sentences were each encoded with an NSE module.
The outputs were combined and passed through a softmax classifier layer to output probabilities.
For SA, we kept the same architecture but used a single NSE layer to encode the input text.

\end{document}